\documentclass{llncs}
\usepackage[utf8]{inputenc}

\usepackage{microtype}
\usepackage{booktabs}
\usepackage{amsmath}
\usepackage{tikz}
\usetikzlibrary{matrix,backgrounds}
\usepackage{tikz-qtree}
\usepackage[vlined,ruled,linesnumbered]{algorithm2e}
\usepackage{array}
\usepackage{cleveref}
\begin{document}

\mainmatter

\title{Hash-based Tree Similarity and Simplification in Genetic Programming for Symbolic Regression}\footnotetext{The final publicaiton is available at \url{https://link.springer.com/chapter/10.1007\%2F978-3-030-45093-9\_44}}

\author{Bogdan Burlacu\inst{1,2} \and Lukas Kammerer\inst{1,2}\\ Michael~Affenzeller\inst{2,3} \and Gabriel~Kronberger\inst{1,2}}
\authorrunning{Burlacu et al.}


\institute{
    Josef Ressel Centre for Symbolic Regression\\
    \and
    Heuristic and Evolutionary Algorithms Laboratory\\
    University of Applied Sciences Upper Austria, Softwarepark 11, 4232 Hagenberg\\
    \and 
    Institute for Formal Models and Verification\\
    Johannes Kepler University, Altenbergerstr. 69, 4040 Linz\\
    \email{bogdan.burlacu@fh-hagenberg.at}
}
\maketitle

\begin{abstract}
We introduce in this paper a runtime-efficient tree hashing algorithm for the identification of isomorphic subtrees, with two important applications in genetic programming for symbolic regression: fast, online calculation of population diversity and algebraic simplification of symbolic expression trees. 
Based on this hashing approach, we propose a simple diversity-preservation mechanism with promising results on a collection of symbolic regression benchmark problems. 
\end{abstract}

\section{Introduction}\label{sec:introduction}
Tree isomorphism algorithms play a fundamental role in pattern matching for tree-structured data. We introduce a fast \emph{inexact}\footnote{Inexact due to the possibility of hash collisions causing the algorithm to return the wrong answer. With a reasonable hahs function, collision probability is negligible.} tree matching algorithm that processes rooted, unordered, labeled trees into sequences of integer hash values, such that the same hash value indicates isomorphism.   
%
We define a distance measure between two trees based on the intersection of their corresponding hash value sequences. We are then able to efficiently compute a distance matrix for all trees by hashing each tree exactly once, then cheaply computing pairwise hash sequence intersections in linear time. 


Genetic Programming (GP) for Symbolic Regression (SR) discovers mathematical formulae that best fit a given target function by means of evolving a population of tree-encoded solution candidates. The algorithm performs a guided search of the space of mathematical expressions by iteratively manipulating and evaluating a large number of tree genotypes under the principles of natural selection. 

Dynamic properties such as exploratory or exploitative behaviour and convergence speed have a big influence on GP performance~\cite{Burke2004}. Exploration refers to the ability to probe different areas of the search space, and exploitation refers to the ability to produce improvements in the local neighbourhood of existing solutions. The algorithm's success depends on achieving a good balance between exploration and exploitation over the course of the evolutionary run~\cite{Crepinsek2013}. 

Population diversity (measured either at the structural-genotypic or semantic-phenotypic level) is typically used as an indirect measure of the algorithm's state-of-convergence, following the reasoning that the exploratory phase of the search is characterized by high diversity and the exploitative phase is characterized by low diversity. Premature convergence can occur at both genotype or phenotype levels, leading to a large amount of shared genetic material in the final population and a very concentrated set of behaviours~\cite{Burks2017}.  

Based on tree node hash values, we propose a new diversity measure defined as an individual's average distance to the rest of the population.  
We show how the resulting diversity score associated with each individual can be used to shift the focus of selection towards fit-and-diverse individuals.  


Section~\ref{sec:methodology} describes the methodology in detail. Section~\ref{sec:results} summarizes the empirical results of our diversity-focused algorithmic improvements, and in Section~\ref{sec:conclusion} we offer our final remarks and conclusions.

\section{Methodology}\label{sec:methodology}
In this section we describe the tree hashing algorithm in detail and introduce a tree distance measure based on the number of common hash values between two tree individuals. The method lends itself well to the efficient computation of distance matrices (for the entire population) since each individual needs only be hashed once and pairwise distances can be subsequently computed using only the corresponding hash value sequences. 

An important aspect of our methodology is the ability of the proposed diversity measure to implicitly capture an individual's semantics by hashing the numerical coefficients associated to the tree's leaf nodes. This allows us to differentiate (in terms of tree distance) between individuals with similar structure but different semantics. 

\subsubsection{Tree hashing}\label{subsec:hashing}

The hashing procedure shares some common aspects with the earlier algorithm by Merkle~\cite{Merkle1988} where data blocks represented as leaves in the tree are hashed together in a bottom-up manner.  

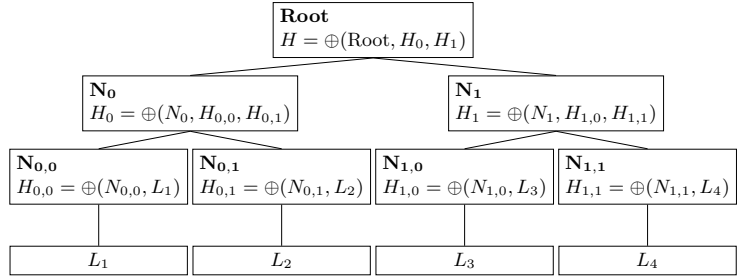
\begin{figure}
    \centering
    \resizebox{0.8\textwidth}{!}{\begin{tikzpicture}
    \tikzset{level distance=12mm}
    \Tree [.\node[draw,align=left]{\textbf{Root}\\$H=\oplus(\mathrm{Root},H_0,H_1)$ }; 
        [.\node[draw,align=left]{$\mathbf{N_0}$\\$H_0 = \oplus(N_0, H_{0,0}, H_{0,1})$};  
            [.\node[draw,align=left]{$\mathbf{N_{0,0}}$\\$H_{0,0} = \oplus(N_{0,0}, L_1)$}; [.\node[draw,align=left,minimum width=29mm]{$L_1$};]] 
            [.\node[draw,align=left]{$\mathbf{N_{0,1}}$\\$H_{0,1} = \oplus(N_{0,1}, L_2)$}; [.\node[draw,align=left,minimum width=29mm]{$L_2$};]] 
        ]                                                                                                         
        [.\node[draw,align=left]{$\mathbf{N_1}$\\$H_1 = \oplus(N_1, H_{1,0}, H_{1,1})$};                   
            [.\node[draw,align=left]{$\mathbf{N_{1,0}}$\\$H_{1,0} = \oplus(N_{1,0}, L_3)$}; [.\node[draw,align=left,minimum width=29mm]{$L_3$};]] 
            [.\node[draw,align=left]{$\mathbf{N_{1,1}}$\\$H_{1,1} = \oplus(N_{1,1}, L_4)$}; [.\node[draw,align=left,minimum width=29mm]{$L_4$};]] 
        ]
    ]
\end{tikzpicture}}
    \caption{Example hash tree where each parent hash value is aggregated from its initial hash value and the child node hash values. Leaf nodes $L_1,...,L_4$ represent data blocks.}\label{fig:hash-tree}
\end{figure}

In our approach an associated data block for each tree node is aggregated together with child node data as input to the hash function to a general-purpose non-cryptographic hash function $\oplus$. This concept is illustrated in~\Cref{fig:hash-tree}, where leaf nodes $L_1,...,L_4$ represent terminal symbols and internal nodes represent mathematical operations. Hash values are computed by the $\oplus$ operator taking as arguments the node's own data hash and its child hash values.  

The procedure given as pseudocode in~\Cref{alg:tree-hashing} relies on the linearization of input tree $T$, such that the resulting array of nodes corresponds to a postorder traversal of $T$. The benefit of linearization is that subtrees are represented by continuous array regions, thus facilitating indexing and sorting operations. For example, tree node $n$ with postorder index $i$ will find its first child at index $j=i-1$, its second child at index $k=j-\mathtt{Size}(j)$ and so on, where $\mathtt{Size}(j)$ returns the size of the subtree whose root node has index $j$.

Sorting the child nodes of commutative symbols is a key part of this procedure. In the general case, this requires a reordering of the corresponding subarrays using an auxiliary buffer. Sorting without an auxiliary buffer is possible when all child nodes are leafs, that is when $\mathtt{Size}(n) = \mathtt{Arity}(n)$ for a parent node $n$. Child order is established according to the calculated hash values. 

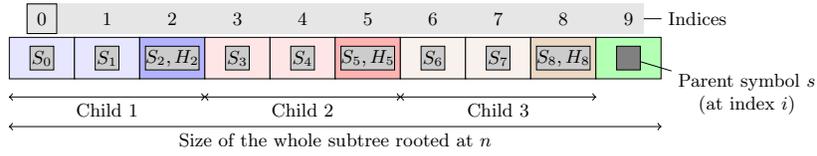
\begin{figure}
    \centering
    \resizebox{0.9\textwidth}{!}{\begin{tikzpicture}[
    array/.style={matrix of nodes,nodes={draw, minimum height=7mm, minimum width=11mm, fill=green!30},column sep=-\pgflinewidth, row sep=0.5mm, nodes in empty cells,
        row 1/.style={nodes={draw=none, fill=none, minimum size=5mm}},
        row 1 column 1/.style={nodes={draw}}}]

        \matrix[array, 
        column 1/.style={nodes={draw, fill=blue!10}},
        column 2/.style={nodes={draw, fill=blue!10}},
        column 3/.style={nodes={draw, fill=blue!30}},
        column 4/.style={nodes={draw, fill=red!10}},
        column 5/.style={nodes={draw, fill=red!10}},
        column 6/.style={nodes={draw, fill=red!30}},
        column 7/.style={nodes={draw, fill=brown!10}},
        column 8/.style={nodes={draw, fill=brown!10}},
        column 9/.style={nodes={draw, fill=brown!30}}
        ] (array) {
            0 & 1 & 2 & 3 & 4 & 5 & 6 & 7 & 8 & 9\\
              &   &   &   &   &   &   &   &   &  \\
        };
\node[draw, fill=gray, minimum size=4mm] at (array-2-10) (box) {};
\node[draw, fill=gray!40, minimum size=4mm, inner sep=0] at (array-2-9) (child3root) { $S_8,H_8$ };
\node[draw, fill=gray!40, minimum size=4mm, inner sep=0] at (array-2-6) (child2root) { $S_5,H_5$ };
\node[draw, fill=gray!40, minimum size=4mm, inner sep=0] at (array-2-3) (child1root) { $S_2,H_2$ };
\node[draw, fill=gray!40, minimum size=4mm, inner sep=0] at (array-2-1) () { $S_0$ };
\node[draw, fill=gray!40, minimum size=4mm, inner sep=0] at (array-2-2) () { $S_1$ };
\node[draw, fill=gray!40, minimum size=4mm, inner sep=0] at (array-2-4) () { $S_3$ };
\node[draw, fill=gray!40, minimum size=4mm, inner sep=0] at (array-2-5) () { $S_4$ };
\node[draw, fill=gray!40, minimum size=4mm, inner sep=0] at (array-2-7) () { $S_6$ };
\node[draw, fill=gray!40, minimum size=4mm, inner sep=0] at (array-2-8) () { $S_7$ };

\begin{scope}[on background layer]
\fill[gray!20] (array-1-1.north west) rectangle (array-1-10.south east);
\end{scope}

\draw[<->]([yshift=-3mm]array-2-1.south west) -- node[below] {Child 1} ([yshift=-3mm]array-2-3.south east);
\draw[<->]([yshift=-3mm]array-2-4.south west) -- node[below] {Child 2} ([yshift=-3mm]array-2-6.south east);
\draw[<->]([yshift=-3mm]array-2-7.south west) -- node[below] {Child 3} ([yshift=-3mm]array-2-9.south east);
\draw[<->]([yshift=-8mm]array-2-1.south west) -- node[below] {Size of the whole subtree rooted at $n$} ([yshift=-8mm]array-2-10.south east);

\draw (array-1-10.east)--++(0:3mm) node [right]{Indices};
\node [align=center, anchor=south, yshift=-7mm, xshift=20mm] at (array-2-10.south) (parent) {Parent symbol $s$\\ (at index $i$)};
\draw (parent)--(box);
\end{tikzpicture}}
    \caption{Example internal node with three child subtrees in postfix representation. Child subarrays are reordered according to priority rules and their respective root hash values $H_2, H_5, H_8$.}\label{fig:sorting-example}
\end{figure}

We illustrate the sorting procedure in \Cref{fig:sorting-example}, where a postfix expression with root symbol $S$ at index 9 contains three child symbols with indices 2, 5, 8 and hash values $H_2, H_5, H_8$, respectively. In order to calculate $\oplus(S)$ the child order is first established according to symbol hash values. Leaf nodes are assigned initial hash values that do not change during the procedure.
Assuming that sorting produces the order $\{ S_5, S_2, S_3 \}$, the child symbols and their respective subarrays need to be reordered in the expression using a temporary buffer. 
After sorting, the original expression becomes:
{\small
\[ \{ [ S_0, S_1, S_2 ], [ S_3, S_4, S_5 ], [ S_6, S_7, S_8 ], S_9 \} \to \{[ S_4, S_5, S_6 ], [ S_1, S_2, S_3 ] , [ S_7, S_7, S_8 ], S_9 \} \]
}
\vspace{-0.5cm}

The sorted child hash values are then aggregated with the parent label in order to produce the parent hash value $H_9 = \oplus(\{H_5, H_2, H_8, S_9\})$. The hashing algorithm alternates hashing and sorting steps as it moves from the bottom level of the tree towards the root node. Finally, each tree node is assigned a hash value and the hash value of the root node is returned as the expression's hash value.

\begin{algorithm}
    \scriptsize
    \caption{Expression hashing}\label{alg:tree-hashing}
    \SetKwInOut{Input}{input}\SetKwInOut{Output}{output}
    \SetKwFunction{Postorder}{Postorder}
    \SetKwFunction{Children}{Children}
    \SetKwFunction{Root}{Root}
    \SetKwFunction{Hash}{Hash}
    \SetKw{Continue}{continue}
    \SetKw{Return}{return}
    \Input{A symbolic expression tree $T$}
    \Output{The corresponding sequence of hash values}
    \BlankLine
    hashes $\gets$ empty list of hash values\;
    nodes $\gets$ post-order list of $T$'s nodes\;
    \ForEach{node $n$ in nodes}
    {
        $H(n) \gets$ an initial hash value\;
        \If{$n$ is a function node}
        {
            \If{$n$ is commutative}{Sort the child nodes of $n$\;\label{alg:expression-hashing-sort}}
            child hashes $\gets$ hash values of $n$'s children\;
            $H(n) \gets \oplus{\left(\textup{child hashes}, H(n) \right)}$\;\label{alg:expression-hashing-compose-hash}
        }
        hashes.append($H(n)$)\;
    }
    \Return hashes\;
\end{algorithm}

\subsubsection{Hash-based tree simplification}\label{subsec:hash-based-simplification}
Our hash-based approach to simplification enables structural transformations based on tree isomorphism and symbolic equivalence relations.
Figure~\ref{fig:simplification-example} illustrates the simplification of an addition function node with two identical child nodes. 

In the expression $E=c_1 x + c_2 yz + c_3 x$ represented in postfix notation, the terms $c_1 x$ and $c_3 x$ are isomorphic and hash to the same value. The simplification algorithm identifies the possibility of folding the constants $c_1$ and $c_3$ so that the two terms are simplified to a single term $c_4 x $ where $c_4 = c_1 + c_3$. The simplified expression then becomes $E'=c_4 x + c_2 yz \equiv E$.

\begin{figure}
    \scriptsize
    \centering
    \begin{tabular}{ccccccccccc}
    $c_3$ & $x$ & $\times$ & $c_2$ & $y$ & $z$ & $\times$ & $c_1$ & $x$ & $\times$ & $+$\\
    \hline
    29375 & 27012 & 29320 & 29375 & 65245 & 52308 & 29320 & 29375 & 27012 & 29320 & 29319\\
\end{tabular}\\
\vspace{5pt}
$\Downarrow$~Sort and hash\\
\vspace{5pt}
\begin{tabular}{ccccccccccc}
    $c_1$ & $x$ & $\mathbf{\times}$ & $c_3$ & $x$ & $\mathbf{\times}$ & $c_2$ & $y$ & $z$ & $\times$ & $+$\\
    \hline
    29375 & 27012 & \textbf{75236} & 29320 & 29375 & \textbf{75236} & 29320 & 65245 & 52308 & 47983 & 31738\\
\end{tabular}\\
\vspace{5pt}
$\Downarrow$~Simplify\\
\vspace{5pt}
\begin{tabular}{cccccccc}
    $c_4$ & $x$ & $\times$ & $c_2$ & $y$ & $z$ & $\times$ & $+$\\
    \hline
    29375 & 27012 & 75236 & 29320 & 65245 & 52308 & 47983 & 31040\\
\end{tabular}

    \caption{Expression simplification. The original expression (top) is sorted and hashed according to Algorithm~\ref{alg:tree-hashing}. Isomorphic subtrees with the same hash value are simplified.}\label{fig:simplification-example}
\end{figure}
\vspace{-1cm}

\subsubsection{Hash-based population diversity}\label{subsec:hash-based-diversity}
Using the algorihm described in Section~\ref{subsec:hashing}, we convert each tree individual into a sequence of hash values corresponding to a post-order traversal of its structure. After this conversion, the distance between two trees can be defined using the intersection between the two sequences\footnote{The S{\o}rensen-Dice coefficient (Equation~\ref{eq:sorensen-dice}) returns a value in the interval $[0,1]$}:
\begin{equation}
    D(T_1, T_2) = \frac{2 \cdot |H_1 \cap H_2|}{|H_1| + |H_2|}\label{eq:sorensen-dice}
\end{equation}
where $H_1, H_2$ represent the hash value sequences of $T_1$ and $T_2$, respectively.

Computing the distance matrix for the entire population can be further optimized by hashing all trees in an initial pass, then using the resulting hash value sequences for the calculation of pairwise distances. 
This leads to the following algorithmic steps:
\begin{enumerate}
    \item Convert every tree $T_i$ in the population to hash value sequence $H_i$ 
    \item Sort each hash value sequence $H_i$ in ascending order\label{step:sorting}
    \item For every pair $(H_i, H_j)$ compute distance using Equation~\ref{eq:sorensen-dice}. 
\end{enumerate}
Sorting in step~\ref{step:sorting} allows us to efficiently compute $|H_i \cap H_j|$ in linear time. As shown in~\ref{tab:distance-performance}, in this particular scenario, these optimizations lead to a two order of magnitude improvement in runtime performance over similar methods~\cite{Valiente2001}. 
\begin{table}
    \centering
    \begin{tabular}{lrr}
        Tree distance method & Elapsed time (s) & Speed-up\\
        \midrule
        Bottom-up                & $1225.751$ & $1.0x$\\
        Hash-based               & $3.677$    & $333.3x$\\
    \end{tabular}
    \caption{Elapsed time computing average distance for 5000 trees}\label{tab:distance-performance}
\end{table}
\vspace{-1cm}

\subsubsection{Diversity as an explicit search objective}\label{subsec:diversity-search-objective}
Diversity maintenance strategies have been shown to improve GP performance~\cite{Crepinsek2013,Burks2017}. A plethora of diversity measures have already been studied: history-based, distance-based, difference-based, entropy-based, etc.~\cite{Crepinsek2013}. However, due to high computational requirements, tree distances like the tree edit distance have seldomly been used. Our hash-based approach overcomes this limitation making it feasible to compute the population distance matrix every generation.

We define an individual's diversity score as its average distance to the rest of the population. The score is easily computed by averaging the corresponding distance matrix row for a given individual. In effect, this causes selection to favor individuals ``farther away'' from the crowd, reducing the effects of local optima as attractors in the search space.

We integrate this new objective into a single-objective approach using standard GP and a multi-objective approach using the NSGA-2 algorithm~\cite{Deb2002}.

In the single-objective case, we extent the standard fitness function with an additional diversity term $d$, such that the new fitness used during selection becomes: $f' = f + d$. Since fitness is normalized between $[0,1]$ both terms have the same scale and no additional weighting is used.

In the multi-objective case the diversity score is used as a secondary objective along with fitness. The NSGA-2 algorithm performs selection using crowding distance within Pareto fronts of solutions.

\section{Empirical Results}\label{sec:results}

We test the proposed approach on a collection of synthetic symbolic regression benchmarks: Vladislavleva~\cite{Vladislavleva2009}, Poly-10~\cite{Poli2003}, Spatial Coevolution~\cite{Pagie1997}, Friedman~\cite{Friedman1991} and Breiman~\cite{Breiman1984}. 
We configure all algorithms to evolve a population of 1000 individuals over 500 generations with a function set consisting of $(+,-,\times,\div,\exp,\log,\sin,\cos,\mathrm{square})$. Different types of mutation (remove branch, replace branch, change node type, one-point mutation) are applied with a probability of 25\%. Tree individuals are initialized using the Probabilistic Tree Creator (PTC2)~\cite{Luke2000}. 

\begin{figure}[ht]
    \centering
    \includegraphics[width=0.49\textwidth]{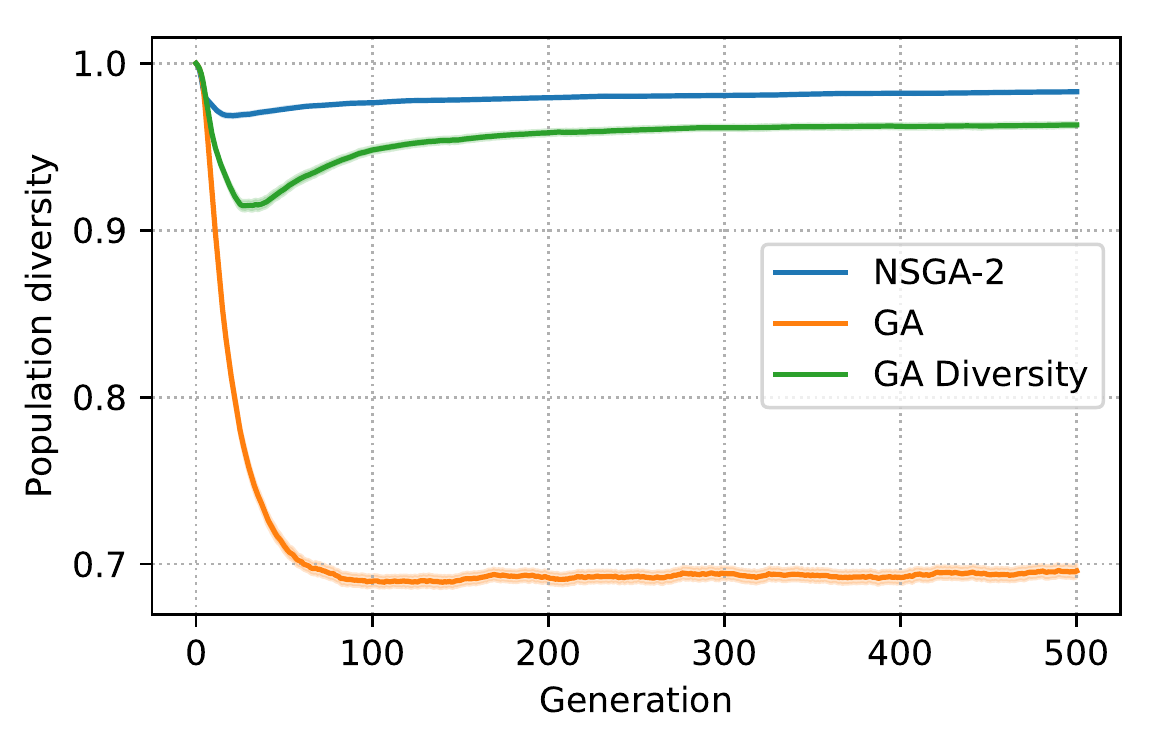}
    \includegraphics[width=0.49\textwidth]{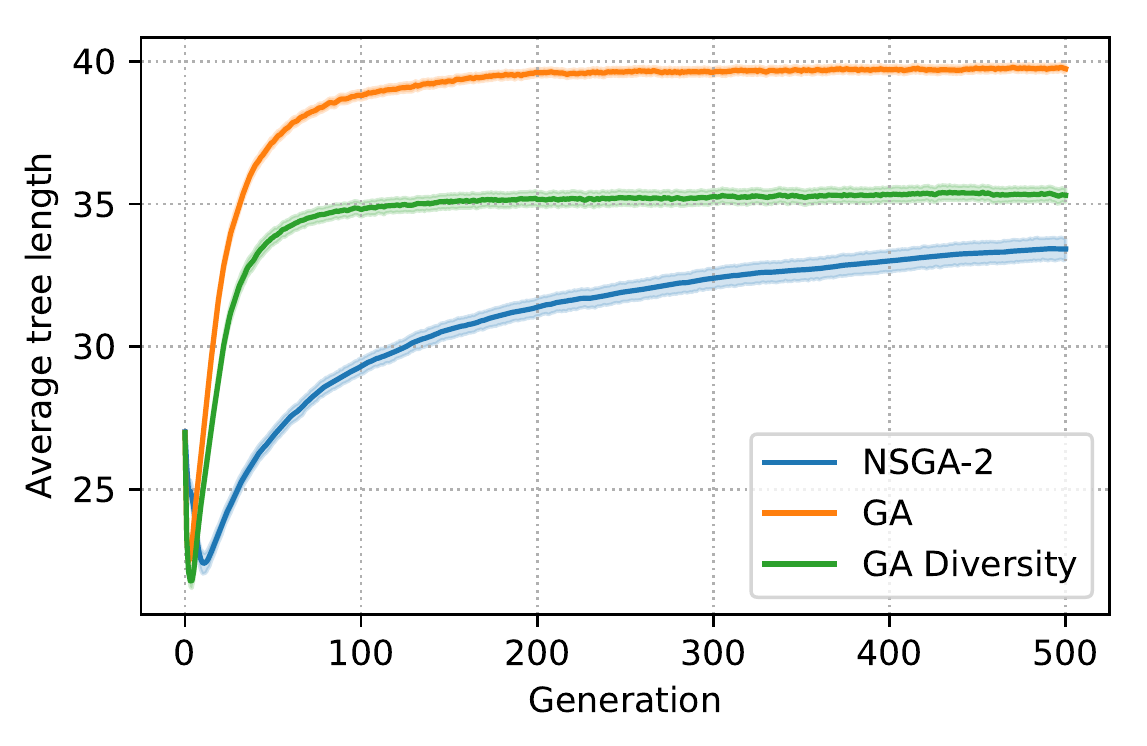}
    \caption{Evolution of average diversity and average tree length}\label{fig:tree-length-similarity}
\end{figure}

The results summarized in Table~\ref{tab:results-summary} show the benefits of diversity maintenance. Selecting for diversity enables both the GA and NSGA-2 algorithms to better exploit population diversity and achieve better results in comparison with the standard GA approach.
The NSGA-2 algorithm in particular is able to avoid overfitting and produce more generalizable models on the Vladislavleva-6 and Vladislavleva-8 problems. In all tested problem instances, both GA Diversity and NSGA-2 outperform standard GA on both training and test data. 
Figure~\ref{fig:tree-length-similarity} shows that GA Diversity and NSGA-2 are able to maintain higher diversity and promote smaller, less bloated individuals. 

The overhead incurred by tree distance calculation depends on the size of the training data. For large data, this overhead becomes negligible as the algorithm will spend most of its runtime evaluating fitness. In our tests GA Diversity is approximately 20-25\% slower than Standard GA. 
A direct runtime comparison with NSGA-2 is not possible due to different algorithmic dynamics.

%

\begin{table}[ht]
    \centering
    \resizebox{0.85\textwidth}{!}{\begin{tabular}{m{2.5cm}m{2.3cm}p{2.3cm}p{2.3cm}r}
Problem data &                   Algorithm &  $R^2$ (training) &  $R^2$ (test) & Time (s) \\
 \midrule
     Breiman - I & GA           & 0.870 $\pm$ 0.040 & 0.865 $\pm$ 0.037 & 1084.5 \\
     Breiman - I & GA Diversity & 0.875 $\pm$ 0.034 & 0.870 $\pm$ 0.029 & 1207.5 \\
     Breiman - I & NSGA-2       & $\mathbf{0.885 \pm 0.013}$ & $\mathbf{0.879 \pm 0.012}$ & 1584.0 \\
 \midrule
    Friedman - I & GA           & 0.859 $\pm$ 0.007 & 0.860 $\pm$ 0.006 & 1090.5 \\
    Friedman - I & GA Diversity & 0.860 $\pm$ 0.007 & 0.860 $\pm$ 0.005 & 1207.0 \\
    Friedman - I & NSGA-2       & $\mathbf{0.863 \pm 0.002}$ & $\mathbf{0.863 \pm 0.003}$ & 1554.0 \\
 \midrule
   Friedman - II & GA           & 0.944 $\pm$ 0.038 & 0.942 $\pm$ 0.041 & 1092.0 \\
   Friedman - II & GA Diversity & 0.957 $\pm$ 0.023 & 0.957 $\pm$ 0.025 & 1248.0 \\
   Friedman - II & NSGA-2       & $\mathbf{0.958 \pm 0.008}$ & $\mathbf{0.957 \pm 0.010}$ & 1474.0 \\
 \midrule
         Pagie-1 & GA           & 0.990 $\pm$ 0.012 & 0.889 $\pm$ 0.107 & 442.5 \\
         Pagie-1 & GA Diversity & 0.994 $\pm$ 0.005 & 0.912 $\pm$ 0.070 & 540.5 \\
         Pagie-1 & NSGA-2       & $\mathbf{0.998 \pm 0.002}$ & $\mathbf{0.932 \pm 0.075}$ & 880.0 \\
 \midrule
         Poly-10 & GA           & 0.820 $\pm$ 0.293 & 0.764 $\pm$ 0.478 & 405.0 \\
         Poly-10 & GA Diversity & 0.840 $\pm$ 0.089 & 0.838 $\pm$ 0.120 & 518.0 \\
         Poly-10 & NSGA-2       & $\mathbf{0.879 \pm 0.072}$ & $\mathbf{0.850 \pm 0.099}$ & 902.5 \\
 \midrule
 Vladislavleva-1 & GA           & 0.999 $\pm$ 0.001 & 0.946 $\pm$ 0.120 & 371.5 \\
 Vladislavleva-1 & GA Diversity & 0.999 $\pm$ 0.001 & 0.980 $\pm$ 0.106 & 450.0 \\
 Vladislavleva-1 & NSGA-2       & $\mathbf{1.000 \pm 0.000}$ & $\mathbf{0.987 \pm 0.026}$ & 784.5 \\
 \midrule
 Vladislavleva-2 & GA           & 0.995 $\pm$ 0.012 & 0.987 $\pm$ 0.028 & 355.0 \\
 Vladislavleva-2 & GA Diversity & 0.995 $\pm$ 0.012 & 0.992 $\pm$ 0.016 & 462.0 \\
 Vladislavleva-2 & NSGA-2       & $\mathbf{0.999 \pm 0.001}$ & $\mathbf{0.998 \pm 0.002}$ & 802.0 \\
 \midrule
 Vladislavleva-3 & GA           & 0.968 $\pm$ 0.062 & 0.932 $\pm$ 0.508 & 445.0 \\
 Vladislavleva-3 & GA Diversity & 0.979 $\pm$ 0.048 & 0.975 $\pm$ 0.053 & 570.0 \\
 Vladislavleva-3 & NSGA-2       & $\mathbf{0.995 \pm 0.014}$ & $\mathbf{0.989 \pm 0.042}$ & 920.5 \\
 \midrule
 Vladislavleva-4 & GA           & 0.951 $\pm$ 0.036 & 0.918 $\pm$ 0.053 & 527.0 \\
 Vladislavleva-4 & GA Diversity & 0.968 $\pm$ 0.023 & 0.936 $\pm$ 0.049 & 651.0 \\
 Vladislavleva-4 & NSGA-2       & $\mathbf{0.982 \pm 0.017}$ & $\mathbf{0.966 \pm 0.027}$ & 973.5 \\
 \midrule
 Vladislavleva-5 & GA           & 0.997 $\pm$ 0.012 & 0.933 $\pm$ 0.144 & 407.0 \\
 Vladislavleva-5 & GA Diversity & 0.999 $\pm$ 0.002 & 0.995 $\pm$ 0.015 & 502.5 \\
 Vladislavleva-5 & NSGA-2       & $\mathbf{1.000 \pm 0.000}$ & $\mathbf{0.997 \pm 0.023}$ & 871.0 \\
 \midrule
 Vladislavleva-6 & GA           & 0.869 $\pm$ 0.155 & 0.072 $\pm$ 0.329 & 369.0 \\
 Vladislavleva-6 & GA Diversity & 0.939 $\pm$ 0.143 & 0.255 $\pm$ 0.978 & 491.0 \\
 Vladislavleva-6 & NSGA-2       & $\mathbf{1.000 \pm 0.023}$ & $\mathbf{1.000 \pm 0.347}$ & 930.0 \\
 \midrule
 Vladislavleva-7 & GA           & 0.895 $\pm$ 0.048 & 0.878 $\pm$ 0.100 & 395.0 \\
 Vladislavleva-7 & GA Diversity & 0.904 $\pm$ 0.029 & 0.892 $\pm$ 0.058 & 501.5 \\
 Vladislavleva-7 & NSGA-2       & $\mathbf{0.917 \pm 0.017}$ & $\mathbf{0.899 \pm 0.033}$ & 843.5 \\
 \midrule
 Vladislavleva-8 & GA           & 0.962 $\pm$ 0.080 & 0.541 $\pm$ 0.569 & 362.5 \\
 Vladislavleva-8 & GA Diversity & 0.986 $\pm$ 0.033 & 0.787 $\pm$ 0.436 & 489.5 \\
 Vladislavleva-8 & NSGA-2       & $\mathbf{0.992 \pm 0.012}$ & $\mathbf{0.817 \pm 0.412}$ & 799.0 \\
\end{tabular}
}
    \caption{Empirical results expressed as median $R^2 \pm$ interquartile range over 50 algorithm repetitions}\label{tab:results-summary}
\end{table}

\vspace{-1cm}

%
%
%

\section{Summary}\label{sec:conclusion}

We described a hashing algorithm for GP trees with applications to distance calculation and expression simplification. 
The approach is highly efficient, making it feasible to measure diversity on a generational basis, as after an initial preprocessing step tree distance is reduced to a simple co-ocurrence count between sorted hash value sequences. With this information new diversity preservation strategies become possible at low computational cost.   

A simple strategy illustrated in this work is to bias selection towards individuals that are more distant from the rest of the population. Experimental results using the standard GA and NSGA-2 algorithms showed increased model accuracy on all tested problem instances, correlated with increased diversity and lower average tree size. Although easily integrated with all GA flavours, we conclude empirically that the strategy is more effective in the multi-objective case when diversity is separately considered. 

Future work in this area will focus on mining common subtrees in the population based on hash value frequencies and designing more complex diversity preservation strategies.

\subsection*{Acknowledgement}
The authors gratefully acknowledge support by the Christian Doppler Research Association and the Federal Ministry of Digital and Economic Affairs within the \emph{Josef Ressel Centre for Symbolic Regression}.

\bibliography{burlacu.bib}
\bibliographystyle{splncs}

\end{document}